\pdfoutput=1

\documentclass[11pt]{article}

\usepackage{acl}

\usepackage{times}
\usepackage{latexsym}
\usepackage{multirow}
\usepackage{graphicx}
\usepackage{soul}  
\usepackage{pifont} 
\usepackage{tikz} 
\usepackage{enumitem}
\usepackage[skins]{tcolorbox}

\setlist{nosep}

\usepackage[T1]{fontenc}

\usepackage{natbib}

\usepackage[utf8]{inputenc}

\usepackage{microtype}
\usepackage{titlesec}  

\newcommand{\bracket}[1]{$\langle$#1$\rangle$}
\newcommand{\cmark}{\ding{51}}%
\newcommand{\xmark}{\ding{55}}%
\newcommand{\dialogmodel}{{\sc {llm-d}}}

\newtcbox{\mybox}[1][green]{on line,
arc=1pt,colback=#1!15!white,colframe=#1!50!black!50!white,
boxsep=0pt,left=1pt,right=1pt,top=2pt,bottom=2pt,
boxrule=1pt}

\newcommand{\shrug}[1][]{%
\begin{tikzpicture}[baseline,x=0.8\ht\strutbox,y=0.8\ht\strutbox,line width=0.125ex,#1]
\def\arm{(-2.5,0.95) to (-2,0.95) (-1.9,1) to (-1.5,0) (-1.35,0) to (-0.8,0)};
\draw \arm;
\draw[xscale=-1] \arm;
\def\headpart{(0.6,0) arc[start angle=-40, end angle=40,x radius=0.6,y radius=0.8]};
\draw \headpart;
\draw[xscale=-1] \headpart;
\def\eye{(-0.075,0.15) .. controls (0.02,0) .. (0.075,-0.15)};
\draw[shift={(-0.3,0.8)}] \eye;
\draw[shift={(0,0.85)}] \eye;
\draw (-0.1,0.2) to [out=15,in=-100] (0.4,0.95); 
\end{tikzpicture}}

\title{Flexible text generation for counterfactual fairness probing}

\author{Zee Fryer\thanks{\ \ Work done as a Google AI Resident.} \quad Vera Axelrod \quad Ben Packer \quad Alex Beutel \quad Jilin Chen \quad Kellie Webster \\
Google Research \\
{\{zeef, vaxelrod, bpacker, alexbeutel, jilinc, websterk\}@google.com}
}

\begin{document}
\maketitle
\begin{abstract}
A common approach for testing fairness issues in text-based classifiers is through the use of counterfactuals: does the classifier output change if a sensitive attribute in the input is changed? Existing counterfactual generation methods typically rely on wordlists or templates, producing simple counterfactuals that don't take into account grammar, context, or subtle sensitive attribute references, and could miss issues that the wordlist creators had not considered. In this paper, we introduce a task for generating counterfactuals that overcomes these shortcomings, and demonstrate how large language models (LLMs) can be leveraged to make progress on this task. We show that this LLM-based method can produce complex counterfactuals that existing methods cannot, comparing the performance of various counterfactual generation methods on the Civil Comments dataset and showing their value in evaluating a toxicity classifier.
\end{abstract}

\section{Introduction}\label{sec:intro}

It is well known that classifiers (such as toxicity detectors) can pick up negative associations about marginalized groups from their training data, e.g. due to under-representation of those groups in the training data, or the higher levels of toxicity in the text data referring to these groups \cite{sap-etal-2019-risk, dixon2018measuring, zhou-etal-2021-challenges}.

One common method of testing classifier models for these unwanted associations is by comparing the classifier's outputs on a particular type of counterfactual text pair: specifically, text pairs which are as similar as possible in format and meaning, but such that one text references a particular sensitive attribute and the other does not (Figure~\ref{fig:intro llm example}; here the sensitive attribute is Islam). If the classifier exhibits a large number of ``flips'' (changes in prediction from original to counterfactual) on these pairs, this indicates a potential problem that may be addressed through mitigations such as
dataset augmentation \cite{dixon2018measuring} or
counterfactual logit pairing \cite{garg-logit-pairing}.




\begin{figure} \small
    \centering
\begin{tcolorbox}[tile, width=\linewidth, rounded corners, arc=5pt, colback=green!5!white, colbacklower=blue!5!white, boxsep=0pt, top=8pt, bottom=8pt, left=10pt, right=10pt]
{\bf Original}: True and the same goes with  \mybox[green]{headscarves}. Its not  religious requirement  but a cultural choice. Simple otherwise there would be no \mybox[green]{Muslim woman} that don't wear them and clearly there are.
\tcblower
{\bf Counterfactual}: True and the same goes with \mybox[blue]{yarmulkes}. Its not \mybox[blue]{a} religious requirement but a cultural choice. Simple otherwise there would be no \mybox[blue]{Jewish man} that don't wear them and clearly there are.
\end{tcolorbox}
\caption{Counterfactual generated by our LLM-based method, given the original text and the prompt ``make this not about Muslims''.}
\label{fig:intro llm example}
\end{figure}

Here we focus on counterfactual generation, and specifically the following questions: 1) How can we efficiently generate large datasets of counterfactual pairs? 2) While preserving the diversity, fluency and complexity of real-world inputs? 3) To probe for subtle or previously-unknown issues?

One approach is to ask humans to create counterfactual counterparts by editing existing examples, but this can be both costly and slow (see e.g. \S3 in \citet{Kaushik2020Learning}). Another method is to use human-curated wordlists to generate counterfactuals: for example to apply ablation or substitution on existing texts or to fill in preset templates \cite{garg-logit-pairing, dixon2018measuring, rudinger2018gender, sheng2019woman}. While these approaches can efficiently generate large datasets (excluding the time required to create the initial wordlists), the results often fail to be fluent, diverse or complex (as we show in Section~\ref{sec:results}) and are not likely to uncover novel issues that the wordlist creators had not considered. 

We suspect that as it becomes more common to use large language models (LLMs) as the base for classifier models (such as toxicity classifiers), these classifiers will become more sensitive to factors such as fluency, word order, and context, and counterfactual generation methods will need to evolve correspondingly to keep up. 

With this in mind, we define a new counterfactual generation task (Section~\ref{sec:task definition}) and demonstrate the potential of existing LLM techniques to address this problem (Section~\ref{sec:llm generation}). Specifically, we show how ideas from \citet{reif2021recipe} can be used to generate natural, diverse, and complex counterfactuals from real-world text examples (as in Figure~\ref{fig:intro llm example}) and combine this with both automated and human evaluation methods (Section~\ref{sec:evaluation contributions}) to ensure that the resulting counterfactuals are of high quality and suited to the task at hand. This human-in-the-loop component also helps to mitigate the risks introduced by using an LLM to generate the text (Section~\ref{sec:generation safety}). Finally, we compare the performance of our method with existing counterfactual generation methods  (Section~\ref{sec:results}), and show that existing methods may not capture certain subtle issues in toxicity classifiers, and that our method addresses some of these deficiencies (Section~\ref{sec:tox detection results}).

We use toxicity detection as a testbed in this work, and focus on generating counterfactuals to probe for false positives -- that is, non-toxic text which is misclassified as toxic due to identity references. While we focus on this particular application to demonstrate one way in which our framework can be useful, it could also be applied in other contexts: for example, probing for false negatives, applications other than toxicity detection, and counterfactual perturbations other than removing the presence of a sensitive attribute.



\section{Related Work}\label{sec:related work}

\subsection{Counterfactual generation}\label{sec:generation related work}

Two common types of counterfactual text pair generation are 1) rule-based methods using templates and/or wordlists, and 2) controlled text generation using deep learning-based language models.

Template-based counterfactual datasets are often built from short, simple sentences: for example, the Jigsaw Sentence Templates dataset
consists of templates such as ``I am a \bracket{adjective} \bracket{identity-label}'' and ``I hate \bracket{identity-label}''.\footnote{\url{https://github.com/conversationai/unintended-ml-bias-analysis}} Other examples include \citet{rudinger2018gender, sheng2019woman}. While this approach provides fine-grained control over identity references and toxicity balance, it also has disadvantages: for example, the resulting text is often not natural and looks quite different from the actual task data. Works such as \citet{prabhakaran-etal-2019-perturbation} and \citet{hutchinson-etal-2020-social} partially mitigate this by using real-world data and targeting specific syntactic slots for substitution, but this can yield incoherent or contradictory text when there are multiple entities referenced in a sentence.
Finally, recent works with templates such as \citet{rottger-etal-2021-hatecheck} and \citet{kirk-hatemoji} have been effective at detailing problems with modern toxicity classifiers, by investing significant targeted effort into probing task-specific functionality, and employing human validation for generated examples.

There have also been attempts to use deep learning to build more general-purpose counterfactual generators. One example is Polyjuice \cite{wu2021polyjuice}, which combines a finetuned GPT-2 model with control codes to generate diverse natural perturbations. But, as we show below, it is difficult to use Polyjuice to modify references to a pre-specified topic. Another example is CAT-Gen \cite{wang2020cat}, which trains an RNN-based encoder-decoder model, using a separate attribute classifier to guide the decoder towards the modifying the desired attribute.  However, both of these require large training sets labeled by sentence permutation type (Polyjuice) or attribute (CAT-Gen).

Other methods combine a pretrained language model with a task-specific classifier, e.g. \citet{Dathathri2020Plug} and \citet{madaan2020generate} which both leverage Bayes' rule to guide text generation while avoiding the need to retrain or finetune the language model itself, and \citet{mostafazadeh-davani-etal-2021-improving} which uses GPT-2 to generate text and uses wordlists and likelihood thresholds to identify valid counterfactuals. 
ToxiGen \cite{toxigen} uses GPT-3 with and without an adversarial classifier-in-the-loop method to generate a large set of challenging examples for toxicity detection, employing identity-specific engineered prompts.
Our method is most similar to these approaches, though we rely less on task-specific classifiers and use generic prompts.

{\renewcommand{\arraystretch}{1.3}  \setlength{\tabcolsep}{5pt}
\begin{table*}\small
\begin{tabular}{l p{0.42\linewidth}p{0.42\linewidth}}
{\bf Attribute} & {\bf Original}  & {\bf \dialogmodel\ rewrite} \\ \hline \hline
LGBQ+ &
How is ``embracing and accepting'' their {\em homosexuality} not a lifestyle choice? & 
How is ``embracing and accepting'' their {\bf love} not a lifestyle choice? \\ \hline

transgender & 
Some people are born {\em transgender}. That appears to be a verifiable fact.  Why is this a question of ``left'' or ``right''? & 
Some people are born {\bf left-handed}. That appears to be a verifiable fact.  Why is this a question of ``left'' or ``right''? \\ \hline

Judaism & 
Get {\em JPFO} up here. If anyone has anything to say about guns it is that organization.  For those that do not know. {\em JPFO is Jews for the Preservation of Firearms}. & 
Get {\bf the NRA} up here. If anyone has anything to say about guns it is that organization.  For those that do not know. {\bf NRA is for National Rifle Association}. \\ \hline

Islam & 
If its {\em Muslim} he's all over it.... I can't figure this guy's loyalty. Who is influencing this guy..... Is it the {\em Muslim Brotherhood, Saudi Arabia, Qatar}?? &
If it's {\bf American} he's all over it.... I can't figure this guy's loyalty. Who is influencing this guy..... Is it the {\bf Democrats, Republicans, Supreme court}?? \\\hline
\end{tabular}
\caption{Examples of \dialogmodel-generated counterfactuals, demonstrating \dialogmodel's ability to make neutral context-aware substitutions or multiple consistent substitutions to remove explicit and implicit references.}
\label{table:llm examples}
\end{table*}}

\subsection{Counterfactual evaluation}\label{sec:evaluation related work}

While most counterfactual generation work includes a definition of what constitutes a ``good'' counterfactual and some method of measuring success relative to these desiderata, the definitions and methods vary depending on factors such as the intended downstream use of the counterfactuals.

Many methods prize counterfactuals with minimal edits relative to the original text and measure success using distance, e.g. \citet{ross2021explaining} \citet{madaan2020generate}. However, this is not well suited for evaluating counterfactuals generated from longer or complex original texts, as these often require multiple edits to remove all references to the sensitive attribute. 
Some methods reward grammaticality but do not require the text to make semantic sense \cite{sheng2020towards}, while others require both fluency and consistency \cite{ross2021tailor, reif2021recipe, madaan2020generate}; some use automated metrics such as perplexity \cite{wang2020cat} and masked language model loss \cite{ross2021explaining} while others use human raters to evaluate fluency \cite{reif2021recipe, wu2021polyjuice}. 


Building on these prior results, we combine several automated metrics to filter out poor quality counterfactuals (e.g. ones with large additions/deletions beyond those required to remove the sensitive attribute).
%
We also develop a human evaluation framework to rate the quality of the counterfactuals that pass automated filtering, with a view to making it easy for human annotators to rate examples quickly and consistently while also rewarding diverse and non-obvious counterfactual generation (e.g. rows 1 and 2 of Table~\ref{table:llm examples}).



\subsection{Toxicity detection}\label{sec:toxicity detection related work}

It is well documented \cite{davidson-etal-2019-racial, dixon2018measuring} that toxicity and hate speech classifiers often pick up on correlations (that are not causations) between references to certain identities and toxic speech: that is, these models incorrectly learn that sensitive attributes such as certain sexual orientations, gender identities, races, religions, etc. are themselves indications of toxicity.

Recent work has gone further and explored the effect of {\em indirect} toxic examples on classifiers \cite{sap-etal-2020-social, lees-etal-2021-capturing, han-tsvetkov-2020-fortifying}, finding that many datasets do not adequately represent this form of toxicity \cite{breitfeller-etal-2019-finding}
and that classifiers are ineffective at identifying it \cite{han-tsvetkov-2020-fortifying}. Based on this, we conjecture that  toxicity classifiers may also associate {\em indirect} references to sensitive attributes with toxicity, which is consistent with \cite{toxigen}. We focus on  exploring this facet of counterfactual probing. 


\section{Methodology} 


Our goal is to detect when a model produces a different score for two examples (original and counterfactual) that differ only by changing a sensitive attribute and that have the same groundtruth label. Ideally the dataset of counterfactual pairs used in this testing should be both large in size and diverse in topic in order to maximise the chances of identifying issues with the model, including issues that the dataset creators may not have considered.

\subsection{Task Definition}\label{sec:task definition}

We define our task as follows:

{\em Given a corpus of text examples that reference a specific sensitive attribute (e.g. a particular religion, LGBQ+ identity, transgender identity), generate a counterfactual text for each original text that preserves both the original label and the original meaning (as far as possible) while removing all references to the chosen sensitive attribute. 

Taken as a set, the counterfactuals should be:
\begin{itemize}
\item {\bf Complex}: The texts should reflect the complexity of expected real-world inputs.
\item {\bf Diverse}: The counterfactual edits should cover a range of topics, both within the attribute's category (e.g. replacing one religion with another) and more generally (replacing specific references with neutral words such as ``person'', ``religion'', etc).
\item {\bf Fluent and consistent}: The generated text should match the style and phrasing of the input text, should be internally consistent (e.g. no changing topic part way through), and should read like plausible natural language.
\end{itemize}}

\subsection{Counterfactual Generation with LLMs}\label{sec:llm generation}

To generate our counterfactuals we build on the results of \citet{reif2021recipe}, which accomplishes a wide range of style transfers using a Transformer-based large language model combined with prompting. Inputs to the LLM consist of three parts: a small fixed set of prompts that demonstrate the rewriting task, the piece of text to be rewritten, and an instruction such as ``make this more descriptive'' or ``make this include a metaphor''. The LLM returns up to 16 attempts at rewriting the input text according to the given instruction.

In order to use this method for counterfactual generation, we retain the prompts used in \citet{reif2021recipe} (see Table~\ref{table:llm prompts} for the full prompt text) but replace the style transfer instruction with ones specific to our task, e.g. ``make this not about Muslims'' or ``make this not about transgender people'' (see Appendix~\ref{appendix:prompt selection} for details). This is one of the few parts of our pipeline that uses the sensitive attribute, and this generalises easily to other attributes simply by changing the instruction.


We use LaMDA \cite{thoppilan2022lamda} as the underlying LLM for text generation in this paper, which belongs to the family of decoder-only Transformer-based dialog models. The LaMDA model used here, which we refer to as \dialogmodel, is described in \S6 of \citet{thoppilan2022lamda}: it has 137B parameters, and was pretrained as a general language model (GLM) on 1.97B public web documents and  finetuned into a dialog model on an additional dataset of curated dialog examples.


For the experiments reported here, we exclusively used the finetuned dialog model: both for safety reasons (\dialogmodel's finetuning includes a focus on reducing toxic text generation \cite{thoppilan2022lamda}) and technical reasons (it could generate longer passages of text than other models available to us). However, we also achieved success using our method with the underlying GLM model (referred to as ``PT'' in \citet{thoppilan2022lamda}), and since prompting techniques have achieved success on multiple different language models \cite{reif2021recipe, brown2020language} we expect that our method would generalise to other LLMs.

\subsection{Counterfactual Evaluation}\label{sec:evaluation contributions}

We evaluate counterfactuals in two phases: an automated phase using a combination of standard metrics and a simple two-layer classifier, and a human evaluation phase based on criteria we developed for rating complex counterfactuals. A key consideration here is that while counterfactuals should be as similar as possible to the originals, they must also remove sensitive attribute references; thus we cannot be \textit{too} strict in enforcing similarity, especially via automated methods.

\dialogmodel \ was configured to generate up to 16 responses for each input, so we use a combination of automated metrics to identify potential good counterfactuals to pass to human raters. In addition to some simple filtering rules (e.g. to catch examples where \dialogmodel \ simply regurgitates its prompt) we use three main metrics:
\begin{itemize}
\item BLEU score \cite{papineni-etal-2002-bleu}, 
\item BERTScore \cite{Zhang2020BERTScore}, and
\item a prediction of whether the sensitive attribute is still referenced (described below).
\end{itemize}
A high BLEU score relative to the original text indicates high lexical similarity \cite[Appendix B]{reif2021recipe}, while a high BERTScore indicates semantic similarity; based on early (separate) tuning experiments, we found that requiring both scores to be above 0.5 was a good trade-off between producing plausible counterfactuals while also allowing some diversity of responses. 

The sensitive attribute predictor is a two-layer fully connected classifier trained for this purpose; full training details are given in Appendix~\ref{appendix:automated metrics}. 
We imposed a threshold of 0.5 on this classifier as well, although the results in this paper suggest that this would benefit from further tuning.

Our human evaluation criteria evaluate the (original, counterfactual) pair along four axes:
\begin{enumerate}
\item fluency/consistency, 
\item presence of sensitive attribute,
\item similarity of label, and 
\item similarity of meaning.
\end{enumerate}
Raters are asked whether the proposed counterfactual is fluent and consistent (yes/no/unsure), whether it references the sensitive attribute (explicitly/implicitly/not at all), whether it should be assigned the same label as the original (yes/no/unsure),\footnote{Note that this criteria is task-dependent; in our case the labels were toxic/nontoxic.} and whether it is similar in meaning and format to the original (scale of 0 to 4).  The full rater instructions are given in Appendix~\ref{appendix:human rater guidelines main header}.  


We use majority vote to consolidate annotator ratings for each example, discarding ties. For our purposes, a counterfactual is deemed ``good'' if it is fluent, does not reference the sensitive attribute, has the same label as the original, and scores at least 2 (out of 4) on similarity of meaning. Thus examples where the majority vote resulted in a rating of ``unsure'' were treated as if they had been rated ``no'' when reporting results in Section~\ref{sec:results}.

\paragraph{Quantifying ``similarity of meaning''}
``Similarity of meaning'' was the hardest metric to define, since removing references to the sensitive attribute often required major edits to the input text. Thus, our score buckets split the counterfactuals in a way that captures both type and severity of edit. This allows us to identify a more diverse pool of good counterfactuals, while also making it easy for users to select a stricter subset if required. 

A score of 4 indicates a perfect ablation counterfactual with no unnecessary changes or new information, 3 means that the counterfactual contains substitutions to similar or neutral words (e.g. ``Muslim'' $\rightarrow$ ``Christian'', ``Judaism'' $\rightarrow$ ``religion''; useful for comparing classifier predictions among identities within a category), while 2 allows for more diverse edits such as minor additions/deletions or substitutions to other topics (useful for initial fairness probing of a model). 1 indicates an example that is reasonably similar to the original but too different to be a useful counterfactual, and a 0 indicates that the text is changed beyond recognition.
See Appendix~\ref{appendix:human rater guidelines main header} for full guidelines and examples.

{\renewcommand{\arraystretch}{1.2}\setlength\tabcolsep{4pt}
\begin{table*} \small
\begin{tabular}{l|p{0.73\linewidth}|cccc}
{\bf Method} & {\bf Text} & {\bf F} & {\bf A} & {\bf L} & {\bf M} \\ \hline\hline
{\em Original} & {\em So you are saying it's OK? Not enough? Long way to go? An apologist for Islamic terrorism? }& - & -  & - & -  \\
Ablation & so you are saying it's ok? not enough? long way to go? an apologist for {\bf terrorism}? & \cmark & \cmark &   \cmark & 4  \\
Substitution & so you are saying it's ok? not enough? long way to go? an apologist for {\bf christian} terrorism? & \cmark & \cmark  & \cmark &  3\\ 
Polyjuice & So you are saying it's OK? Not enough? Long way to go? An apologist for {\bf secular enlightenment}? & \cmark& \cmark  & \cmark &  2\\ 
\dialogmodel & So you are saying it's OK? Not enough? Long way to go? An apologist for {\bf terrorism}? & \cmark & \cmark & \cmark &  4  \\  \hline

{\em Original} & {\em Shocking that this article didn't once mention ``islam'', ``islamic'' or ``Muslim''. } & - & - & - & -\\
Ablation & shocking that this article didn't once mention {\bf ``''}, {\bf ``''} or {\bf ``''}. & \xmark & \cmark & -  & - \\
Substitution & shocking that this article didn't once mention ``{\bf christianity''}, ``{\bf christian}'' or ``{\bf christian}''.  & \cmark & \cmark  & \cmark & 3\\ 
Polyjuice & {\bf Waving} that this article didn't once mention ``islam'', ``islamic'' or ``{\bf islandic}''. & \xmark  & \xmark & - &  - \\ 
\dialogmodel & Shocking that this article didn't once mention ``{\bf christian}'', ``{\bf christians}'' or ``{\bf Christ}''. & \cmark & \cmark & \cmark & 3 \\ 
 \end{tabular}
\caption{Civil Comments examples referencing Islam, with generated counterfactuals and human annotations ({\bf F}luent, doesn't reference {\bf A}ttribute, similarity of {\bf L}abel, and similarity of {\bf M}eaning). In the top example all methods produced a reasonable counterfactual; in the second, only \dialogmodel\ and substitution generated good counterfactuals.}
\label{table:side by side examples by method}
\end{table*}}

\subsection{Safety}\label{sec:generation safety}

Large language models come with safety and toxicity issues \cite{bender2021stochasticparrots, abid2021large}, which is of particular concern when using them to generate text for the purpose of counterfactual fairness probing in other models. The \dialogmodel \ model has been finetuned by its creators to help mitigate some of these safety concerns \cite[\S6]{thoppilan2022lamda}, and we also built safeguards into our pipeline to reduce the chances of our method producing problematic or toxic counterfactuals. Even with human-in-the-loop, it is still possible for our method to produce some problematic examples, e.g. ones that perpetuate negative stereotypes, but we aim to reduce this risk.

First, we only aim to generate counterfactuals in the sensitive $\rightarrow$ neutral direction. That is, we choose input texts that reference the sensitive attribute, and ask \dialogmodel \ to remove these references; we do NOT ask the model to generate text about marginalized groups starting from neutral texts (though in practice it can sometimes substitute one identity group for another). Additionally, our evaluation setup ensures that all generated text is checked by at least one human, specifically includes a criteria checking for {\em implicit} references to the identity as well as explicit ones, and includes a ``reject for other reason'' box to allow raters to remove examples if either the original or counterfactual text contains negative stereotypes or hate speech. This provides a second line of defence against any toxic text that might slip through.

\section{Implementation}

\subsection{Data} \label{sec:data methodology}
The CivilComments dataset \cite{DBLP:journals/corr/abs-1903-04561} is a set of approximately 2 million English-language internet comments from 2015-2017, with crowdsourced toxicity annotations. CivilComments-Identities (CC-I) is a 450k subset of CivilComments where each text has additional crowdsourced labels for references to various identities, such as gender,\footnote{The available labels in this category are {\tt male}, {\tt female}, {\tt transgender}, and {\tt other\_gender}, which should always be used with extreme care to avoid any implication that ``male''/``female'' and ``transgender'' refer to disjoint sets (see e.g. \citet{larson-2017-gender}); for this work we use only the {\tt transgender} label.} 
sexual orientation, religion, and race.  

Our experiments focus on four identity subcategories in CC-I, namely {\tt muslim}, {\tt jewish}, {\tt transgender}, and {\tt homosexual\_gay\_and\_lesbian},\footnote{Despite the name, this category appears to have been more broadly interpreted by crowd raters as referring to all non-straight sexual orientations, and therefore we continued to treat it as such in our experiments.} which for simplicity we refer to as {\tt LGBQ+}. These categories were chosen because they are all groups that have faced high levels of online toxicity that may have bled through into classifier models (e.g. \citet{abid2021large, DBLP:journals/corr/abs-1903-04561}), and because the annotators we used for our experiments were sufficiently familiar with these categories to evaluate our generated counterfactuals effectively.

We restrict our dataset to texts between 10 and 45 words long that do not contain URLs, for ease of analysis by human raters.
We further require that texts have a score of at least 0.8 for the relevant attribute, and a toxicity score of at most 0.1: i.e. least 80\% of the CC-I annotators agreed that the text referenced the specified attribute/identity, and at most 10\% of them viewed the comment as toxic. 

We chose to focus on only non-toxic examples (as rated by the CC-I annotators) in our experiments, because toxic examples can introduce an unwanted confounding factor: there are many examples in the dataset that are only toxic because they contain a slur, and removing or substituting the slur often renders the resulting text non-toxic. Since we are focused on the ability to generate counterfactuals with the same label as the original, we excluded these examples from our dataset. Note that this a choice we make in the context of this particular application, but the general methodology described here could also be used to investigate toxic original examples if deemed appropriate.


\subsection{Counterfactual generation}\label{generation methodology}

We compare our \dialogmodel-based generation method to three other methods: ablation, substitution, and the Polyjuice counterfactual generator \cite{wu2021polyjuice}. We summarise each of these methods here, and full details are given in Appendix~\ref{appendix:generation methods experimental setup header}.

We generate a list of keywords relevant to each topic using frequency analysis on the entire \mbox{CC-I} corpus, followed by manual curation to remove words that often co-occur with a sensitive attribute but are not specific to that topic (e.g. ``discriminating'' and ``surgery'' for transgender identity).

To generate ablation counterfactuals, we replace any occurrence of the keywords in our input examples with the empty string. For substitution, we replace all religion-based keywords with a corresponding concept from Christianity, and all sexuality/gender words with their ``opposite'', e.g. ``gay'' $\rightarrow$ ``straight'', ``transgender'' $\rightarrow$ ``cisgender''. Keywords with no obvious replacement (e.g. ``transition'', or ``Israel'') are left unchanged. Note that this can make the substitution method appear artificially good at performing multiple consistent substitutions within a sentence, something that can usually only be achieved with complex rule-based systems (e.g. \citet{lohia2022counterfactual}), and comes at the cost of limited counterfactual diversity. This is discussed further in Section~\ref{sec:results} when comparing the results of substitution and \dialogmodel\ counterfactuals.

{\renewcommand{\arraystretch}{1.2}\setlength\tabcolsep{7pt}
\begin{table*} \small \centering
\begin{tabular}{lr|rrr|rrr}
\multirow{2}{*}{Method} & \multirow{2}{*}{\# examples}& \multirow{2}{*}{Fluent} &\multirow{2}{*}{Attribute ref} & \multirow{2}{*}{Label} & \multicolumn{3}{c}{FAL and...} \\
  & & & & & Meaning 4 &  Meaning 3+ & {\bf Meaning 2+} \\ \hline \hline
Ablation & 200 & 46.6 & 87.6 & 99.5 & 33.0 &  33.0   &  33.0 \\
Substitution & 200 & 96.5 & 88.7 & 100.0 &  0.0 & 84.0  &  84.5 \\
Polyjuice & 162 & 71.2 & 15.4 & 88.8 & 2.5  & 4.9 &  10.5 \\
\dialogmodel & 191 &  95.7 & 71.1 & 96.3 &  14.1 &  39.3 & 62.3 \\
\end{tabular}
\caption{Percentage of counterfactuals (generated from Islam-referencing texts) that were labeled by annotators as being fluent, not referencing the sensitive attribute, and having the same label as the original, respectively. ``FAL and Meaning $n$+'' lists the percentage of examples that satisfied all of these criteria {\em and} were given a score of $n$ or higher by annotators for similarity of meaning.} 
\label{table:annotator breakdown for methods}
\end{table*}}

In order to fairly compare our \dialogmodel\ method with Polyjuice, we generated 16 Polyjuice counterfactuals per input: 8 with no constraints on generation, and 8 where we first used our ablation keyword list to replace all topic-specific keywords in the input sentence with the token {\tt [BLANK]}. These 16 results were then filtered and ranked in the same way as with \dialogmodel, and the top result returned.

Examples for each method are given in Table~\ref{table:side by side examples by method}.

\subsection{Counterfactual evaluation}\label{sec:evaluation methodology}

All human annotation of our generated counterfactuals were performed by three of the authors. Each annotator initially rated a subset of the examples, divided to ensure that every counterfactual received at least two ratings, and any examples with non-unanimous scores were passed to the third rater (with scores hidden) for a tiebreaker vote. Examples that received three distinct ratings for a category (yes/unsure/no) were discarded; the only exception to this was the Similarity of Meaning category, where we averaged the raters' scores. 

In order to ensure rating consistency and refine the clarity of the instructions, we performed two smaller rounds of test annotation first (50-100 examples) followed by a review session to discuss examples with divergent scores or ``unsure'' ratings. While the annotators were of diverse genders (male, female, non-binary) and moderately to extremely familiar with the sensitive attributes chosen for our experiments, we also note that they were all white citizens of Western countries and that this could have informed their interpretation of the toxicity task and what substitutions are ``neutral''.  

\subsection{Toxicity detection}\label{sec:toxicity detection methodology}

We use our generated counterfactuals to evaluate the robustness of the Perspective API toxicity classifier to counterfactual perturbations.\footnote{\url{www.perspectiveapi.com}}
 Perspective API defines toxicity as ``a rude, disrespectful, or unreasonable comment that is likely to make you leave a discussion''; the toxicity score  is the predicted probability of a reader perceiving the input as toxic.

We focus on the change in predicted toxicity score from original to counterfactual. 
This is both because any toxicity cut-off threshold will likely vary by use-case, and because we expect that large changes in score will provide interesting and useful information about the classifier even if they do not happen to straddle the toxicity threshold.

\section{Results}\label{sec:results}

{\renewcommand{\arraystretch}{1.2}\setlength\tabcolsep{7pt}
\begin{table*} \small \centering
\begin{tabular}{llr|rrr|rr}
\multirow{2}{*}{Method} & \multirow{2}{*}{Topic} & \multirow{2}{*}{\# examples}& \multirow{2}{*}{Fluent} &\multirow{2}{*}{Attribute ref} & \multirow{2}{*}{Label} & \multicolumn{2}{c}{FAL and...} \\
  & & & & & &   Meaning 3+ & {\bf Meaning 2+} \\ \hline \hline
\multirow{4}{*}{\dialogmodel}  & LGBQ+ & 99 &  100.0 & 54.6 & 98.6 & 24.2 & 48.5\\
 & transgender & 99 & 97.9 & 43.9 & 98.5 & 22.2 & 36.4\\ 
 & Judaism & 95 & 96.7& 58.4 & 96.2 & 41.1 & 50.5\\
 & Islam & 94 & 95.6 & 67.0 & 97.5 & 35.1 & 58.5\\ \hline

\multirow{4}{*}{substitution}  & LGBQ+ & 20 &  93.8 & 92.3 & 100.0 & 50.0 & 50.0\\
 & transgender & 20 & 95.0 & 42.1 & 100.0 & 35.0 & 35.0\\
 & Judaism & 20 & 89.5 & 57.9& 100.0 & 50.0 & 50.0\\
 & Islam & 20 & 100.0 & 80.0 & 100.0 & 80.0 & 80.0\\
\end{tabular}
\caption{Percentage of examples satisfying each rating criteria, split by topic. Columns are similar to Table~\ref{table:annotator breakdown for methods}.}
\label{table: annotator breakdown for topics}
\end{table*}}

\subsection{Comparison of generation methods}\label{sec:compare gen methods results}
We sample 200 examples that reference Islam from our curated subset of CivilComments-Identities and generate a counterfactual with each of four methods: ablation, substitution, Polyjuice, and our \dialogmodel-based method. The resulting 753 counterfactuals were shuffled and split between the three annotators for rating;\footnote{Neither \dialogmodel \ nor Polyjuice always successfully generated valid counterfactuals, resulting in 191 \dialogmodel\ counterfactuals and 162 Polyjuice counterfactuals.} annotators had access to the sensitive attribute label but not the generation method for each example. The results are given in Table~\ref{table:annotator breakdown for methods}. Recall that for our purposes, a counterfactual is ``good'' if it is fluent, does not reference the sensitive attribute, has the same label as the original, and scores at least 2 on similarity of meaning (bolded column in Table~\ref{table:annotator breakdown for methods}).

In Table~\ref{table:annotator breakdown for methods}  
we see that ablation counterfactuals are often not fluent, but that when the input text can be ablated successfully (e.g. sentences where the keywords are used as adjectives, such as ``The Muslim woman...'') the resulting counterfactuals all receive the maximum score for Similarity of Meaning. Polyjuice was generally unsuccessful at removing references to the sensitive attribute, despite the use of {\tt [BLANK]} tokens to direct the model to the portions of the sentence requiring editing.
While substitution achieved higher success rates than \dialogmodel \ in this experiment, we show in Section~\ref{sec:compare topics results} below that this may partly have been due to the choice of topic and/or wordlist; this breakdown also does not capture the diversity of topics in the generated counterfactuals.

Finally, we note that the subset of input texts for which ablation produced a good counterfactual tended to be the ``easy'' examples, in that substitution produced a good counterfactual for 98.5\% of this subset, and \dialogmodel \  75\%. 

\subsection{Generation on multiple topics}\label{sec:compare topics results}

We sample 100 examples from our curated subset of CivilComments-Identities for each of the attributes Judaism, LGBQ+, and transgender, along with a subset of 100 examples referencing Islam from the set used above. 
Annotators had access to the sensitive attribute label for each example while rating. Results are given in Table~\ref{table: annotator breakdown for topics}.

The key observation here is that our \dialogmodel-based method generalises easily to multiple topics. We also see further evidence (e.g. attribute reference in Table~\ref{table: annotator breakdown for topics}) that our pipeline's automated ranking requires further finetuning, in particular for identifying counterfactuals which have successfully removed all references to the sensitive attribute. In examining discarded \dialogmodel \  responses we found that of the 200 examples where the top-ranked \dialogmodel \ response did not pass human rating, 105 of these examples (52.5\%) had a plausible counterfactual further down the ranking (as judged by one annotator); including these in our evaluation would have raised \dialogmodel's overall success rate to 75\%. 

We also generate substitution counterfactuals for a subset of 20 randomly selected examples for each topic, and find that substitution performs almost identically to \dialogmodel \ at generating good counterfactuals for each of the non-Islam topics. For the LGBQ+ and transgender categories in particular, this may be due to the fact that explicit labels are most commonly used only to refer to minority groups: one talks about ``same-sex marriage'' and ``transgender athletes'', but simply ``marriage'' and ``athletes'' when referring to the majority group. Thus a reference to e.g. ``cisgender atheletes'' still carries an implicit reference to transgender issues. This highlights the need for more complex and diverse counterfactual generation techniques that do not rely solely on substitutions and wordlists.

\subsection{Toxicity detection}\label{sec:tox detection results}
Throughout this section we restrict our attention only to the ``good'' counterfactuals (as rated by the human annotators) because poor-quality ones can produce artificially high or low swings in toxicity (due to changing the text too much relative to the original, or by failing to remove the sensitive attribute, respectively); we omit Polyjuice because it produced too few good examples to analyse.

\begin{figure}
    \centering
    \includegraphics[width=0.90\linewidth]{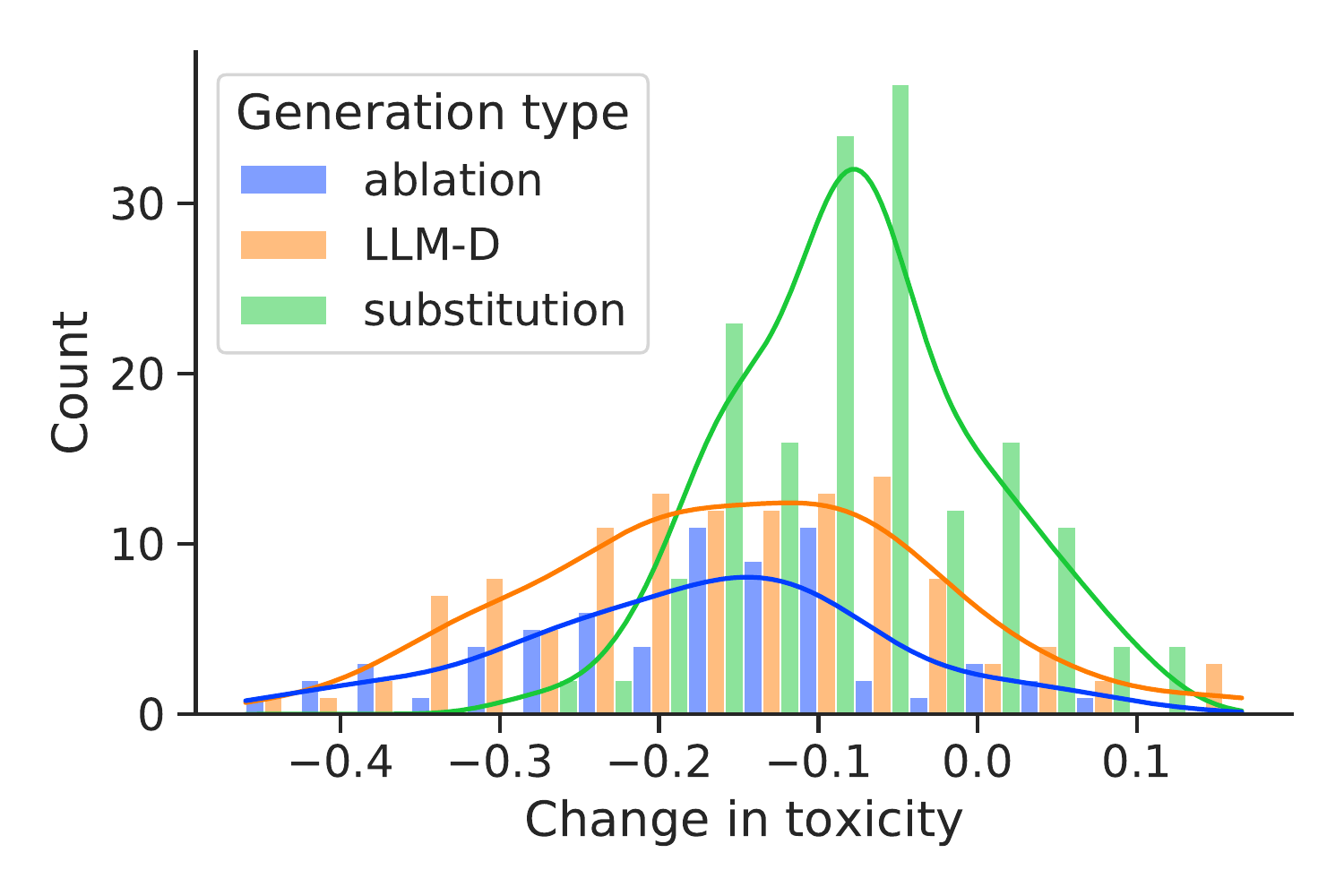}
    \caption{Differences in toxicity score from original texts to their counterfactuals; negative scores indicate that Perspective API rated the counterfactual {\em less likely} to be viewed as toxic than the original.}
    \label{fig:muslim toxdiff hist}
\end{figure}

Counterfactuals generated by all methods have lower predicted toxicity scores on average than the original Islam-referencing texts, as shown in Figure~\ref{fig:muslim toxdiff hist}; see also Figure~\ref{fig:muslim toxdiff scatter by gen type} in the appendix for a more detailed breakdown. Substitution produce the smallest change in toxicity scores: an average difference of -0.08, compared to -0.15 for \dialogmodel\ and -0.17 for ablation. This suggests that counterfactuals generated by \dialogmodel \ and other methods may be producing more challenging examples for the classifier than substitution is, possibly because substitution (by design!) produces text that stays within the same broad topic, and this lack of diversity can make it harder to uncover unexpected negative associations in the classifier. 


We also look at the average change in toxicity score across the four topics 
for both \dialogmodel \  and substitution-generated counterfactuals (Table~\ref{table: avg toxicity diff by topic}). While the sample sizes are too small to draw  concrete conclusions, the small average change in toxicity for religion-referencing substitution counterfactuals compared both to other topics and to \dialogmodel-generated counterfactuals reinforces the conjecture that the toxicity classifier may view all references to religion as similarly toxic. This suggests that more diverse counterfactuals are indeed necessary to effectively probe a model for subtle counterfactual fairness issues. 

Note that the average change in toxicity score is not necessarily meaningful to an end-user. For example, if Perspective API is used to  
remove comments online with scores above a certain threshold, only score changes around that threshold will have a noticeable end-user impact.
Figures~\ref{fig:muslim toxdiff scatter by gen type} and \ref{fig:topics toxdiff scatter by topic} in the appendix provide a more detailed breakdown of how these score changes were distributed, which can help to place the above results in context. However, for the purposes of counterfactual fairness probing we believe it is still important to look at all score changes, not only those near the cut-off point, as this can help to identify areas of potential bias {\em before} end-users are affected.

{\renewcommand{\arraystretch}{1.2} \setlength\tabcolsep{7pt}
\begin{table} \small
    \centering
\begin{tabular}{llrr}
Method & Topic & \# ex & Avg tox diff \\ \hline\hline
\multirow{4}{*}{\dialogmodel}  & LGBQ+ & 48 & -0.25\\
 & transgender & 36 & -0.10 \\ 
 & Judaism & 48 & -0.11 \\
 & Islam  & 55 & -0.15 \\ \hline

\multirow{4}{*}{substitution}  & LGBQ+ & 10 & -0.28\\
 & transgender & 7 & -0.15 \\
 & Judaism & 10 & -0.04 \\
 & Islam & 16 & - 0.05\\
\end{tabular}
    \caption{Average difference in toxicity from original to counterfactual, measured on the good counterfactual pairs generated in Section~\ref{sec:compare topics results}. A negative value indicates that the Perspective API classifier found the counterfactual {\em less} toxic than the original.}
    \label{table: avg toxicity diff by topic}
\end{table}}

\section{Conclusion}\label{sec:conclusion}

\paragraph{Our Contributions}
We have defined a new counterfactual generation task for fairness probing of text classifier models, and have shown that several common types of methods fail to satisfy the requirements of this task and that these failures may limit the effectiveness of the resulting counterfactuals in probing these classifier models. We further show that our \dialogmodel-based approach combined with automated and human rating can generate high quality, diverse, and complex counterfactual pairs from real-world text examples.

\paragraph{Usage and Limitations}

Counterfactuals generated via our \dialogmodel-based approaches could used both to test for undesired behaviour in classifiers and potentially to mitigate that behaviour via methods such as dataset augmentation,
as has been found useful in various settings, e.g. \citet{dinan-etal-2020-queens}, \citet{hall-maudslay-etal-2019-name}.


However, we emphasise that this is not without risk. Language models are known to produce toxic text \cite{wallace-etal-2019-universal} and reflect or amplify biases from their training data \cite{sheng-etal-2019-woman}, among other problems \cite[\S5]{foundations}; we always recommend human review on at least a subset of the data when using potentially sensitive generated text. Language is contextual, and there is a great deal of social context that must be accounted for when attempting to evaluate the behaviours and biases of machine learning models and generated text, so it is important for human review to be performed by a diverse pool of reviewers knowledgeable about the downstream task and the social issues at play (in contrast to the small set of annotators for this illustrative study). 

Using text generated from methods such as ours is also not appropriate in all situations. For example, we emphasise that this generated data should be used to {\em augment} other forms of data, not replace it. 
Similarly, while this study sought to generate diverse texts for analysis, a more restrictive definition of counterfactual may be appropriate when using generated text to {\em mitigate} classifier issues, e.g. by using a stricter cut-off for the ``Similarity of Meaning'' evaluation criteria.

\paragraph{Future Research}
There are several areas of future research to highlight. Most generally, for this investigation we focused on one way this framework can be useful, and made several narrowing choices; however, our framework can be useful in other contexts and applications such as investigating false negatives (by considering original examples that are toxic), probing other types of classifiers than toxicity models, or generating other types of counterfactuals than simply removing the sensitive attribute (e.g. rewording text to explore model robustness). Furthermore, our method would benefit from improved control over the LLM-generated text through e.g. prompt tuning \cite{lester-etal-2021-power}, demonstration-based prompt-engineering and adversarial decoding \cite{toxigen}, or finetuning \cite{wei2021finetuned}, as well as more effective filtering of counterfactuals that still reference the sensitive attribute.



\section*{Acknowledgements}
We would like to thank Vinodkumar Prabhakaran, Ian Tenney, Emily Reif, Ian Kivlichan, Lucy Vasserman, Blake Lemoine, Alyssa Chvasta, and Kathy Meier-Hellstern for helpful discussions and feedback at every stage of this project.

\bibliography{references, custom}

\appendix

\section{\dialogmodel \ counterfactual text generation}\label{appendix:llm main section header}

\subsection{Setup}\label{appendix:llm setup}
Following \citet{reif2021recipe}, we use ``\{'' and ``\}'' delimiters in the formatting of the prompt to encourage \dialogmodel \  to provide its response in a similar format, and automatically discard any text outside of the first set of delimiters in each response. The prompts are formatted in a second-person conversational style, as this is the style of data that \dialogmodel \  was finetuned on; for a template suitable for standard next-token language models, see \cite[Table 7]{reif2021recipe}.

While our initial experiments used the underlying General Language Model (GLM) part of \dialogmodel, all results in this paper were generated using \dialogmodel. We used a temperature of 1 and $k=40$ for the top-$k$ next token sampling, and we did not discard responses regardless of the ``safety'' score \dialogmodel\ assigned to them, as we found that this too severely curtailed the diversity of responses. We mitigated the safety risks of this by ensuring that we had a robust human evaluation step in place later in the pipeline.

We also filter out responses with failure modes observed often in early experiments, including responses that were just a string of punctuation or the ``shrug emoji'' \shrug, verbatim repetitions of the input text, and responses that regurgitated part of the prompt (``here is a rewrite...'', ``here is some text...''). These filters were applied to the initial set of 16 responses from the model.

\subsection{Prompt and instruction selection}\label{appendix:prompt selection}

The full set of prompts used in all of our experiments are listed in Table~\ref{table:llm prompts}; these are the same prompts used in \citet{reif2021recipe}. 

We experimented with different prompts, but found that more task-specific prompts did not produce measurably better results, and in fact found that \dialogmodel \  tended to overfit much more strongly to the final few prompts when the prompts specifically referenced the sensitive attribute. For example, using a set of 7 prompts demonstrating examples of counterfactual generation  specifically for transgender identity, where the last two prompts referenced beauty pageants and Kiwi transgender weightlifter Laurel Hubbard respectively, the (unfiltered) \dialogmodel \  responses to the 100 transgender-referencing examples used in Section~\ref{sec:compare topics results} included 22 references to New Zealand, 31 references to weight lifters, and 5 references to beauty queens / beauty pageants. By comparison, using the prompts in Table~\ref{table:llm prompts} generated 0 results involving any of these keywords, and a total of 7 results referencing bells/snow/trees (see the final two prompts in Table~\ref{table:llm prompts}).

For the rewriting instruction, we found that ``make this not about [sensitive attribute]'' helped to focus the language model's attention on the desired parts of the sentence (as opposed to Polyjuice, which would often produce permutations that were completely unrelated to the sensitive attribute reference in the sentence) but that this did not reliably translate into {\em removing} the reference to the sensitive attribute. However, the fact that \dialogmodel \  produces 16 independent responses meant that there was consistently at least one response that did satisfy the criteria to be a good counterfactual, and one direction of future work is to automatically identify these responses more effectively.


\subsection{Automated metrics}\label{appendix:automated metrics}

We used the implementation of BLEU score provided by the {\tt sacrebleu} package, using the NIST smoothing method as described in \citet{chen-cherry-2014-systematic} to mitigate the fact that we are using a corpus-level metric to compute scores on individual pairs of sentences. 

The implementation of BERTScore is the one provided by the authors \cite{Zhang2020BERTScore}, modified to accept a Flax-based BERT model. BERTScore computes both a recall score (which rewards text pairs where everything in the original sentence is also represented in the counterfactual) and a precision score (rewards pairs where everything in the counterfactual is also represented in the original).\footnote{Note that these are not symmetric: for example, a counterfactual that simply repeats the original text but adds an extra detail to the end would score more highly on recall than precision.} We use the resulting F1 score as our metric since we want our counterfactuals to neither add too much nor delete too much compared to the original text.

The attribute classifier is also JAX/Flax-based, and comprises a 2-layer fully connected network (hidden dimension 2048), using the first token of the input text's BERT representation \cite{devlin-etal-2019-bert} as the embedding function. It was trained on a subset of CivilComments-Identities (all texts, regardless of toxicity, that referenced at least one attribute of interest with a score $ >0.5$, along with 20k negative examples that referenced none of the attributes of interest) using the AdamW optimizer \cite{loshchilov2018decoupled} (with learning rate 0.001, weight decay 0.002) for 36k steps with a batch size 256, using a binary cross-entropy loss function to allow for multi-label predictions.

\section{Counterfactual generation methods}\label{appendix:generation methods experimental setup header}
\subsection{Ablation}\label{appendix:ablation experimental setup}

For ablation, we generate a list of key terms per identity and simply remove those terms from each text. The lists for each attribute are in Table \ref{table:ablation words}.

The term list was generated by fitting a unigram naive bayes classifier to the non-toxic subset of Civil Comments data (toxicity $< 0.1$), separating texts labeled with the given identity group (attribute score $> 0.5$) from a random sample of the rest. The 20 features (unigrams) most strongly associated with the identity class were used as the candidate wordlist, and were filtered by hand to remove irrelevant terms.

We emphasise that these wordlists are {\em not} complete representations of the corresponding attributes and that our ablation counterfactuals were generated purely to provide a baseline score for comparison to other methods.

\subsection{Substitution}\label{appendix:substitution experimental setup}

To generate counterfactuals using substitution, we take the ablation wordlists and (where possible) assign each one a corresponding word from another identity in the same broad category, e.g. replacing one religion with another. For examples with no plausible substitution (e.g. ``transition'' in the transgender category) we leave the word unchanged. See Table~\ref{table:substitution words} for the full set of word pairs.

As with the ablation wordlists above, we emphasise that these are not necessarily complete representations of the corresponding attributes. They were generated purely for the purposes of providing a baseline for comparison in our experiments, and should not be used as-is to generate counterfactuals for fairness probing in real world settings.

\subsection{\dialogmodel}\label{appendix:llm experimental setup}

We use the same fixed set of prompts for every input text (see Appendix~\ref{appendix:prompt selection} and Table~\ref{table:llm prompts}), and the instruction ``make this not about X'', where X is the sensitive attribute referenced in the input text. \dialogmodel \  generates up to 16 responses per input, which are filtered as described in Appendix~\ref{appendix:automated metrics} and then ranked by taking the average of their BLEU score and BERTScore F1 score. Only the top-ranked example is returned for rating.

For some inputs, it can happen that none of \dialogmodel's responses are of sufficient quality to pass the filtering step. We rerun the generation pipeline on each of these failed inputs until a counterfactual is returned, up to a maximum of 5 attempts.

\subsection{Polyjuice}\label{appendix:polyjuice experimental setup}

Polyjuice \cite{wu2021polyjuice} is a general-purpose counterfactual generator that uses a finetuned LM  (GPT-2) along with control-codes to generate diverse permutations of sentences. To our knowledge, Polyjuice has not been evaluated for fairness probing, but its flexible generation abilities make it a promising approach to compare with.

A Polyjuice user can choose from various types of permutation (negation, shuffle, deletion, etc) and can even specify where in the sentence the edit should be made by replacing words or phrases with the {\tt [BLANK]} token.

For each input text, we generate 16 potential counterfactuals: 8 where we allow Polyjuice to choose which parts of the text to modify, and 8 where we direct its attention to the sensitive attribute reference(s) by replacing all keywords from the corresponding ablation list with the {\tt [BLANK]} token. These 16 examples are then filtered and ranked by the same criteria as the \dialogmodel \  examples, and the top-scoring one is returned for rating.

As with the \dialogmodel \  counterfactuals, we rerun the Polyjuice generation pipeline on inputs that failed to return a valid counterfactual, up to 5 times per example.

We found that some of the generated Polyjuice counterfactuals still contained tokens such as {\tt BLANK} or {\tt EMPTY}, which we replaced with the empty string before rating the examples.  

\section{Human rater guidelines}\label{appendix:human rater guidelines main header}

We rate each (original, counterfactual) pair on four criteria:
\begin{itemize}
\item fluency
\item presence of sensitive attribute
\item similarity of meaning
\item similarity of label
\end{itemize}
Precise definitions of each criteria (along with examples) are given in the rater instructions below. For our purposes, we required an (original, counterfactual) pair to be rated ``yes'' for fluency, ``no reference'' for presence of sensitive attribute, ``yes'' for similarity of label, and at least 2 (out of 4) for similarity of meaning. 

These guidelines are designed to be flexible and easily modifiable to apply to multiple text formats (internet comments, video titles, etc) and downstream tasks (toxicity detection, sentiment analysis, etc). We highlight the parts that are specific to our task by \ul{[bracketing and underlining]} them.

\subsection{Rater instructions}\label{appendix:rater instructions}
\begin{enumerate}\setlength\itemsep{0.5em}
\item {\bf {\em Fluency/consistency} \newline
Is the text fluent and consistent, and does it make sense? (yes/no/unsure)}

Ignore any extraneous punctuation marks at the beginning or end of the text. You should also ignore capitalisation / lack thereof unless it makes a significant difference to the fluency.

Evaluate the fluency relative to the original text and in the context of the domain: these examples are \ul{[internet comments]}, so they need not adhere to perfect spelling and grammar to be realistic examples.

Examples of sufficiently fluent sentences:
\begin{itemize}
\item ``islam is great, its just a fact'' [fine: minor incorrect grammar but not unrealistic for internet speech]
\item ``Some folks are born white... just accept it.'' [fine: awkward phrasing but not inconsistent]
\end{itemize}

Examples of sentences that are not consistent or do not make sense:
\begin{itemize}
\item ``{\em Christians} want to impose {\em sharia law} on the country.'' [inconsistent: Sharia law is associated with Islam]
\item ``The problem with on violence is it harms'' [does not make sense]
\item ``---!!!!!--,,,,,,xxxxz'' [gibberish]
\end{itemize}

\item {\bf {\em Presence of sensitive attribute} \newline
Does the text:
\begin{itemize}
\item explicitly reference the attribute?
\item implicitly reference the attribute?
\item not reference the attribute?
\end{itemize}}

The targeted attribute will be specified for each example; references to other sensitive attributes (e.g. references to other religions when the specified attribute is Islam) do not count.

An {\em implicit reference} is a group or identity referenced but not stated explicitly. For example:

\begin{itemize}
\item ``Jack went on a date with his boyfriend'' implicitly references the fact that Jack is queer but does not state it explicitly.
\item ``Our state is trying to pass a bathroom bill, this sucks'' implicitly references transgender identity, since ``bathroom bill'' is commonly understood to be a type of law that targets trans and nonbinary people.
\item ``She wore a hijab on stage'' is an implicit reference to Islam, since the hijab is a common marker of Muslim identity.
\end{itemize}

If the text contains an explicit reference to the attribute, there's no need to evaluate it for implicit references as well.

\item {\bf {\em Similarity of meaning} \newline
Taking into account both the changes made to remove references to the sensitive attribute and any other changes in the sentence, how similar is the text to the original? (scale of 0 to 4)}

\begin{itemize}
\item {\bf Score of 4}: Retains the same grammatical structure and overall meaning as the original, and changes only the parts of the sentence required to remove the sensitive attribute (no other changes or deletions).
\item {\bf Score of 3}: Substitutes or removes the sensitive attribute, only differing from the original in ways that don't materially affect the meaning of the example, e.g. minor grammatical changes, use of synonyms, substitutions to similar or semantically neutral concepts (``BLM'' $\rightarrow$ ``environmental protestors'', ``Allah'' $\rightarrow$ ``God'' are examples of similar substitution, ``Muslims'' $\rightarrow$ ``people'', ``Antifa'' $\rightarrow$ ``protest movement'' are examples of neutral substitution).

Note that here ``neutral'' means words like ``people'' ,``citizens'', ``protest movement'', ``religion'' etc. Non-toxic substitutions that change the meaning of the sentence should be assigned a score of 2 (see below).
\item {\bf Score of 2}: Makes an additional change (addition or removal) beyond the minimal changes required to remove the sensitive attribute, and/or makes edits that change the meaning of the sentence.
\item {\bf Score of 1}: Bears some resemblance to the original sentence, but has major changes beyond those necessary to remove the sensitive attribute.
\item {\bf Score of 0}: Is completely unrelated to the original sentence.
\end{itemize}

Here are some example texts for each score.

\begin{itemize}
\item Original: 
\begin{itemize}
\item How would you feel about opening a meeting with a Muslim prayer to Allah?
\end{itemize}

\item Score of 4:
\begin{itemize}
\item How would you feel about opening a meeting with a prayer? [no changes beyond minimal necessary]
\item How would you feel about opening a meeting with prayers?
\end{itemize}
\item Score of 3:
\begin{itemize}
\item How would you feel about opening a meeting with a {\em Christian} prayer to {\em God}? [substitution to similar concept]
\item How would you feel about opening a meeting with a {\em religious reading}? [substitution to neutral concept]
\item How would you feel about opening {\em this event} with a prayer? [minor extra rewording beyond minimum necessary]
\end{itemize}
\item Score of 2:
\begin{itemize}
\item {\em Since we're in Portland}, how would you feel about opening this meeting with a prayer? [addition of extra detail in addition to necessary chances]
\item How would you feel about opening a meeting with a {\em brief political statement}? [substitution that changes the meaning of the sentence]
\item How would you feel about opening a meeting? [removal of extra detail that changes the meaning in a non-trivial way]
\end{itemize}
\item Score of 1:
\begin{itemize}
\item What's all this about meetings and prayers? [similar topic but major changes to original beyond those necessary to remove attribute]
\end{itemize}
\item Score of 0:
\begin{itemize}
\item By the banks of the Seine. [no resemblance to original]
\end{itemize}
\end{itemize}

\item {\bf {\em Similarity of label}\newline
Should this example be assigned the same ground-truth \ul{[toxicity]} label as the original text? (yes/no/unsure)}

Note: you are not trying to determine whether the two texts are equivalent or whether the concepts referenced in either text are comparable or even true: {\em your answer should depend only on whether you would think it would cause harm if a \ul{[toxicity]} classifier model treated the two texts differently.}

\ul{[For the purposes of this question, we define toxicity to mean ``a rude, disrespectful, or unreasonable comment that is likely to make you leave a discussion''.]}\footnote{This is the definition of toxicity used by Perspective API; \url{https://developers.perspectiveapi.com/s/about-the-api-key-concepts}}

The following are examples where both texts {\em should} be assigned the same label, because a \ul{[toxicity]} classifier model should treat them the same way:

\begin{itemize}
\item Text: Miss Nevada (2021) is a beauty queen. \newline
Original: Miss Nevada (2021) is a transgender woman. \newline
(both non-toxic)

\item Text: Maybe the lead scientist is a woman. \newline
Original: Maybe the lead scientist is gay.  \newline
(both non-toxic)

\item Text: I hate white people. \newline
Original: I hate straight people. \newline
(both toxic)
\end{itemize}

The following are examples where the example should {\em  not} be assigned the same label as the original text, because a \ul{[toxicity]} classifier model should {\em  not} treat them the same way:

\begin{itemize}
\item Text: All white men are dangerous. (toxic) \newline
Original: All white men think cats are great. (non-toxic)

\item Text: Can't trust a velociraptor, they don't have morals. (non-toxic) \newline
Original: Can't trust a Christian, they don't have morals. (toxic)
\end{itemize}

\item {\bf \em {Reject for other reason?} (yes/leave blank)}
Use this field to flag examples that pass all of the other criteria but seem inappropriate for use as counterfactuals.
\end{enumerate}

\subsection{Example}\label{appendix:human rater example}

{\bf Text}: He was a baker? I did not see that in the story. \newline
{\bf Original}: He was Muslim? I did not see that in the story. \newline
{\bf Sensitive attribute}: Islam

\begin{enumerate}\setlength\itemsep{0.5em}
\item Is the text fluent and consistent, and does it make sense? (yes/no/unsure) \newline
{\tt Yes}

\item Does the text:
explicitly reference the attribute?
implicitly reference the attribute?
not reference the attribute? \newline
{\tt Not reference the attribute}

\item Taking into account both the changes made to remove references to the sensitive attribute and any other changes in the sentence, how similar is the text to the original? (scale of 0 to 4) \newline
{\tt 2}

\item Should this example be assigned the same ground-truth toxicity label as the original text? (yes/no/unsure) \newline
{\tt Yes}

\item Reject for other reason? (yes or \bracket{leave blank}) \newline
{\tt \bracket{leave blank}}
\end{enumerate}


{\renewcommand{\arraystretch}{1.2}
\begin{table*}
\centering
\begin{tabular}{p{0.15\linewidth}p{0.7\linewidth}}
{\bf Attribute} &  {\bf Ablation unigrams}\\ \hline \hline
LGBQ+ & gay, gays, homosexual, homosexuality, homosexuals, lesbian, lesbians, lgbt, lgbtq, sexuality\\
transgender & bathroom, bathrooms, hormones, lgbt, lgbtq, testosterone, trans, transgender, transgendered, transition\\
Judaism & holocaust, israel, israeli, israelis, jew, jewish, jews, judaism, semitic, semitism, zionist\\
Islam & allah, hijab, islam, islamic, islamist, islamists, islamophobia, koran, mosque, mosques, muslim, muslims, quran, sharia, sunni\\
\end{tabular}
\caption{Ablation Wordlists, generated as described in Appendix~\ref{appendix:ablation experimental setup}. Note that these are not intended to be comprehensive wordlists for each topic, nor are all of the words direct references to the attribute itself (e.g. ``Israel'' or ``bathroom''); we chose to retain these indirect references if they appeared in the top 20 unigrams produced by the naive Bayes classifier since we were evaluating the resulting counterfactuals on implicit references to the attribute as well as explicit ones. }
\label{table:ablation words}
\end{table*}}

{\renewcommand{\arraystretch}{1.2}
\begin{table*}
\centering
\begin{tabular}{p{0.15\linewidth}p{0.15\linewidth}  p{0.6\linewidth}}
{\bf Attribute} & {\bf Replacement category} & {\bf Substitution wordpairs}\\ \hline \hline
LGBQ+ & heterosexual & (gay, straight), (gays, straights), (homosexual, heterosexual), (homosexuality, heterosexuality), (homosexuals, heterosexuals), (lesbian, straight), (lesbians, straights), (lgbt, straight), (lgbtq, straight)\\
transgender & cisgender & (lgbt, cis), (lgbtq, cis), (trans, cis), (transgender, cisgender), (transgendered, cisgendered)\\
Judaism & Christianity & (jew, christian), (jewish, christian), (jews, christians), (judaism, christianity)\\
Islam & Christianity & (allah, god), (hijab, cross), (islam, christianity), (islamic, christian), (islamist, fundamentalist), (islamists, fundamentalists), (islamophobia, anti-christian bias), (koran, bible), (mosque, church), (mosques, churches), (muslim, christian), (muslims, christians), (quran, bible), (sharia, canon law), (sunni, catholic)\\
\end{tabular}
\caption{Substitution Wordlists. Note that while some of the pairings are direct analogs (e.g. ``gay'' $\rightarrow$ ``straight'', ``Muslim'' $\rightarrow$ ``Christian''), others were chosen to maximise the chances of generating valid counterfactuals while retaining the general meaning of the sentence (e.g. ``LGBTQ'' $\rightarrow$ ``straight''/``cis'', ``hijab'' $\rightarrow$ ``cross''); we are {\em not} implying that all of these pairings are completely analogous.}
\label{table:substitution words}
\end{table*}}

\begin{figure*}
    \centering
    \includegraphics[width=0.9\linewidth]{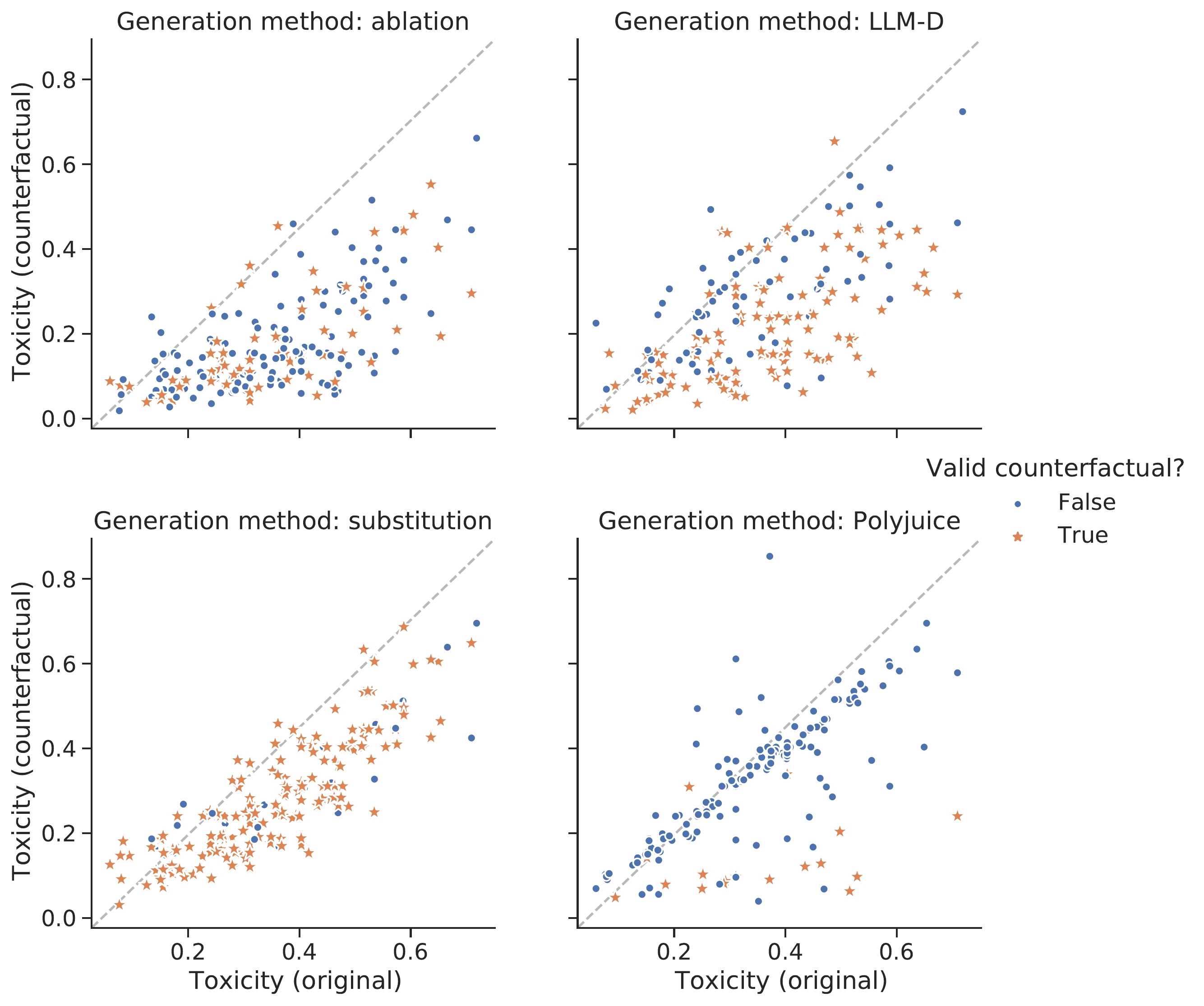}
    \caption{Toxicity scores of counterfactuals (generated from the Islam-referencing texts in Section~\ref{sec:compare gen methods results}) plotted against the toxicity scores of their original text; points in the lower right portion of each graph correspond to examples where Perspective API rated the counterfactual as {\em less} likely to be toxic than the original. We include the counterfactuals that did not pass the human rating step in order to illustrate the effects of different counterfactual generation methods on toxicity detection: for example, ablation failed mostly on the fluency criteria so its ``poor'' counterfactuals still exhibit a drop in toxicity here, whereas Polyjuice failed mostly on removing references to the sensitive attribute so its ``poor'' counterfactuals tend to cluster around the $y=x$ line.}
    \label{fig:muslim toxdiff scatter by gen type}
\end{figure*}

\begin{figure*}
    \centering
    \includegraphics[width=0.9\linewidth]{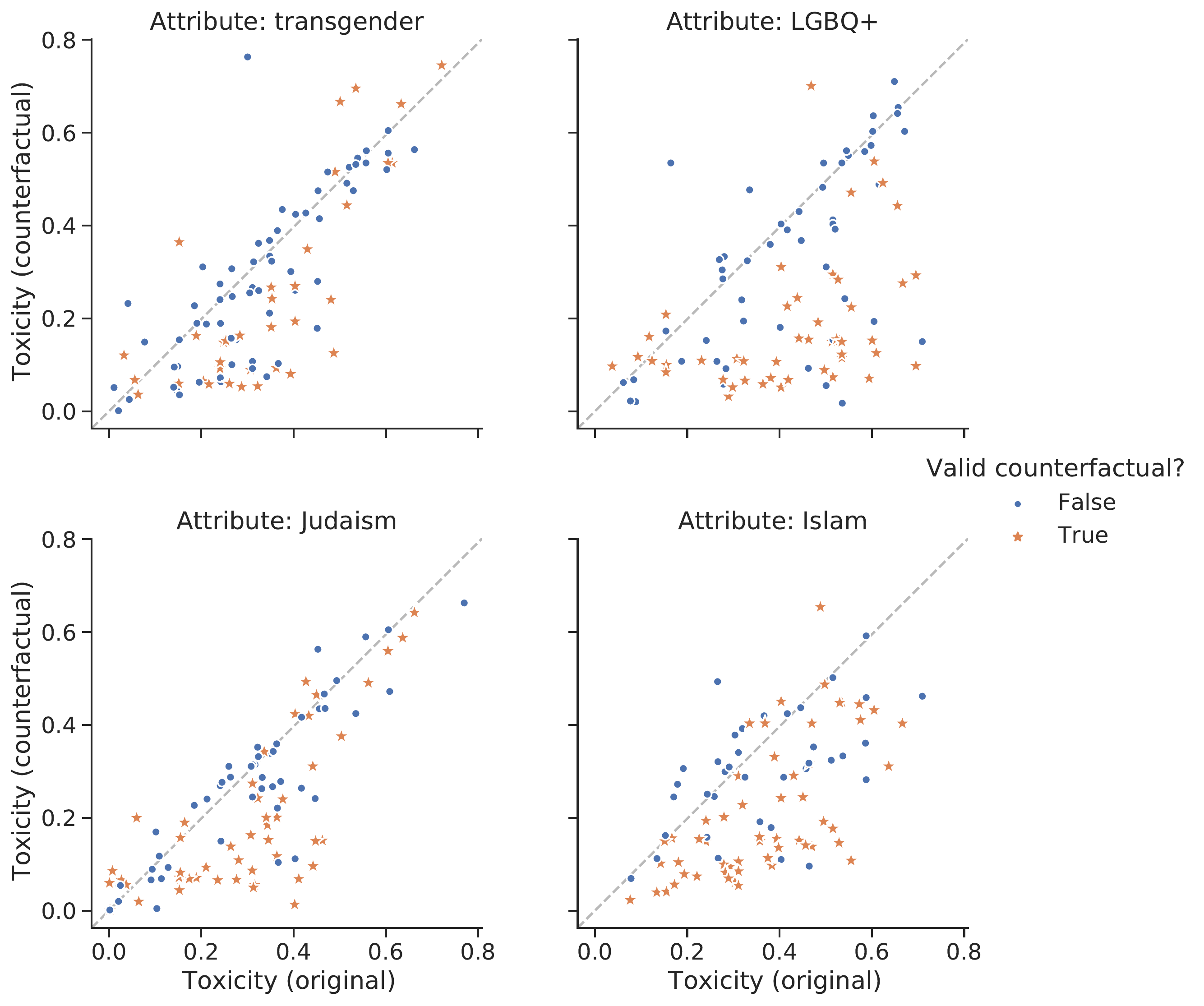}
    \caption{Toxicity scores of \dialogmodel \  counterfactuals (from Section~\ref{sec:compare topics results}) plotted against the toxicity scores of their original text; points in the lower right portion of the graph correspond to examples where Perspective API rated the counterfactual as {\em less} likely to be toxic than the original. Observe that the LGBQ+ category sees some swings in toxicity score that could result in a label flip (represented by the points closest to the bottom right corner), a phenomenon which is much less common with the other three topics.}
    \label{fig:topics toxdiff scatter by topic}
\end{figure*}

\begin{table*}
\newlength{\promptspace}
\setlength{\promptspace}{1em}

{\tt 0: Here is some text: \{When the doctor asked Linda to take the medicine, he smiled and gave her a lollipop.\}. Rewrite it to be more scary.\vspace{\promptspace}

 1: \{When the doctor told Linda to take the medicine, there had been a malicious gleam in her eye that Linda didn't like at all.\} \vspace{\promptspace}
 
 0: Here is some text: \{they asked loudly, over the sound of the train.\}. Rewrite it to be more intense.\vspace{\promptspace}
 
 1: \{they yelled aggressively, over the clanging of the train.\}\vspace{\promptspace}
 
 0: Here is some text: \{When Mohammed left the theatre, it was already dark out\}. Rewrite it to be more about the movie itself.\vspace{\promptspace}
 
 1: \{The movie was longer than Mohammed had expected, and despite the excellent ratings he was a bit disappointed when he left the theatre.\}\vspace{\promptspace}
 
 0: Here is some text: \{next to the path\}. Rewrite it to be about France.\vspace{\promptspace}
 
 1: \{next to la Siene\}\vspace{\promptspace}
 
 0: Here is some text: \{The man stood outside the grocery store, ringing the bell.\}. Rewrite it to be about clowns.\vspace{\promptspace}
 
 1: \{The man stood outside the circus, holding a bunch of balloons.\}\vspace{\promptspace}
 
 0: Here is some text: \{the bell ringing\}. Rewrite it to be more flowery.\vspace{\promptspace}
 
 1: \{the peales of the jangling bell\}\vspace{\promptspace}
 
 0: Here is some text: \{against the tree\}. Rewrite it to be include the word ``snow''.\vspace{\promptspace}
 
 1: \{against the snow-covered bark of the tree\}'\vspace{\promptspace}
 
 0: Here is some text: \{{\color{blue} \bf input text here}\}. Rewrite it to be {\color{blue} \bf instruction here}}.

\caption{The full prompt text used to generate counterfactuals with \dialogmodel. The ``0'' and ``1'' indicate two speakers, since \dialogmodel \  expects inputs formatted in turns of conversation. The text to be rewritten and the corresponding instruction (e.g. ``not about transgender people'') are inserted in the last line (blue, boldface).}
\label{table:llm prompts}
\end{table*}

\end{document}